\documentclass[letterpaper, 10 pt, conference]{ieeeconf}
\pdfoutput=1 % Comment this line out
\IEEEoverridecommandlockouts                              % This command is only
\usepackage[utf8]{inputenc}
\usepackage[T1]{fontenc}
\usepackage{textcomp}
\usepackage{graphicx}
\usepackage{authblk}
\usepackage{amssymb}
\usepackage{biblatex}
\graphicspath{ {./images/} }
\usepackage{hyperref}

\begin{document}
\date{}
\title{\LARGE \bf
Art Style Classification with Self-Trained Ensemble of AutoEncoding Transformations
}

% \author{ \parbox{2 in}{\centering Akshay Joshi 
%          \\
%          {\tt\small s8akjosh@stud.uni-saarland.de}}
%          \hspace*{ 0.2 in}
%          \parbox{2 in}{ \centering Ankit Agrawal
%          \\

%          {\tt\small s8anagra@stud.uni-saarland.de}}
%          \hspace*{ 0.2 in}
%          \parbox{2 in}{ \centering Sushmita Nair
%          \\
%          {\tt\small s8sunair@stud.uni-saarland.de}}
%          \hspace*{ 0.2 in}

% }

\author{\bf{Akshay Joshi}}
\author{\bf{Ankit Agrawal}}
\author{\bf{Sushmita Nair\vspace{-1em}}}
\affil[]{Universität des Saarlandes\vspace{-1em}}
\affil[]{\textit {\{s8akjosh, s8anagra, s8sunair\}@stud.uni-saarland.de\vspace{2em}}}

\maketitle
\thispagestyle{empty}
\pagestyle{empty}

%%%%%%%%%%%%%%%%%%%%%%%%%%%%%%%%%%%%%%%%%%%%%%%%%%%%%%%%%%%%%%%%%%%%%%%%%%%%%%%%
%\begin{abstract}
\section*{\bf{ABSTRACT}}
\label{sec:abstract}

The artistic style of a painting is a rich descriptor that reveals both visual and deep intrinsic knowledge about how an artist uniquely portrays and expresses their creative vision. Accurate categorization of paintings across different artistic movements and styles is critical for large-scale indexing of art databases. However, the automatic extraction and recognition of these highly dense artistic features has received little to no attention in the field of computer vision research. In this paper, we investigate the use of deep self-supervised learning methods to solve the problem of recognizing complex artistic styles with high intra-class and low inter-class variation. Further, we outperform existing approaches by almost 20\% on a highly class imbalanced WikiArt dataset with 27 art categories. To achieve this, we  train the EnAET semi-supervised learning model (Wang et al., 2019) with limited annotated data samples and supplement it with self-supervised representations learned from an ensemble of spatial and non-spatial transformations.

%\end{abstract}

%%%%%%%%%%%%%%%%%%%%%%%%%%%%%%%%%%%%%%%%%%%%%%%%%%%%%%%%%%%%%%%%%%%%%%%%%%%%%%%%
{\bf{Keywords:}} Art Style Recognition, WikiArt, Self-supervised Learning, Feature Extraction\\

\section{\bf{INTRODUCTION}}
\label{sec:introduction}

Art style recognition and classification has been recently gaining interest in the fields of Image Processing and Computer Vision because of the availability of large datasets with visually rich artistic images and paintings. This is due to the fact that art libraries are consistently being digitized and indexed into several categories based on artist, painting style, and genre. Different art styles are easily perceived and comprehended by humans. But, the properties such as sentiment and emotion are difficult to be modeled in a computational way. 

\begin{figure}[htbp]
    \centering
    \includegraphics[scale=0.95]{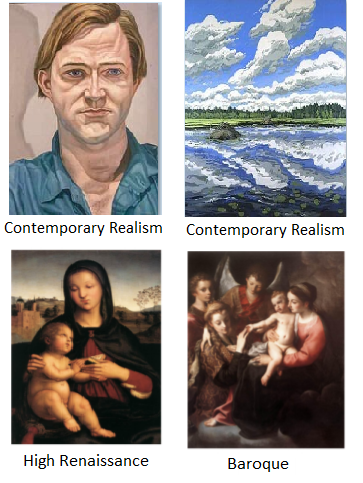}
    \caption{Four examples of images of the WikiArt datasets. The top two images represent the problem of variation found within a style and the bottom two represents similarity between different styles.}
    \label{fig:images_types_examples}
\end{figure}

In this paper, we explore the problem of classifying complex art works and paintings by its style (e.g. Impressionism, Baroque, etc.). This task is considerably challenging as the dataset is highly skewed and has very high intra-class and inter-class variations i.e. there is a variety of distinct painting styles but existing in the same art class, as well as similar painting styles but spread across different art categories. To understand the idea of these challenges, some examples taken from the WikiArt dataset are reported in Figure~\ref{fig:images_types_examples}.

The rest of the paper is organized as follows: In Section \ref{sec:related_works}, we look into related works that have been done in this domain. We discuss about how the data is acquired including some of the augmentations that have been used in Section \ref{sec:data_acq}. Further, in Section \ref{sec:methodology} we describe our approach followed by various baselines we used to assess the performance of our approach in Section \ref{sec:baselines}.
In Section \ref{sec:exp_and_results}, we present experimental results and compare them against the baselines. Finally, we conclude with inferences and future directions.

\section{\bf{RELATED WORK}}
\label{sec:related_works}

A lot of research has been done in the domain of paintings categorization mainly for styles, genres, and artists. In this section, we will discuss some work done in the field of art style recognition and classification.

Several publications have addressed this problem using pre-computed features such as color histograms, special organisation and lines descriptions ({\cite{c1, c2}}), or directly training features from the images itself ({\cite{c3, c4}}). Good results were achieved using pre-computed features using multi-task learning and dictionary learn by (\cite{c2, c4}), addressed the problem using a variation of the same neural network and managed to achieve better results with a fully automatic procedure. More recently, \cite{c5} suggested that the ResNet50 is the best performing CNN model with an accuracy of 62\% on WikiArt data. The results obtained in this study show that self-supervised learning methods outperform other CNN methods used with even less computation and data required compared to other methods.

\section{\bf{DATA ACQUISITION}}
\label{sec:data_acq}

To train our models, we use the Wikipaintings dataset, a large image dataset collected from WikiArt which is made publicly available following the experimental protocol used by Tan et al., 2016. The resulting dataset contains more than 80,000 fine-art paintings for a total of 27 styles: Abstract Art, Abstract Expressionism, Art Informal, Art Nouveau (Modern), Baroque, Color Field Painting, Cubism, Early Renaissance, Expressionism, High Renaissance, Impressionism, Magic Realism, Mannerism (Late Renaissance), Minimalism, Naive Art (Primitivism), Neoclassicism, Northern Renaissance, Pop Art, Post Impressionism, Realism, Rococo, Romanticism, Surrealism, Symbolism, and Ukiyo-e. 

Out of a total of 81,446 images, for the ResNet\cite{c6} baseline setup, we took 40,000 images for training, 20,168 images for validation and 8,616 images as test set. Similarly, for EnAET\cite{c7} setup, we used only a subset of data having a total of 30,183 images, among which 58\% is used to training i.e. 17,567 images, 4000 images for validation, and 29\% for test i.e 8,616 images. We've used the same test set for all models to obtain comparable results.

\begin{figure}[htbp]
    \centering
    \includegraphics[scale=0.65]{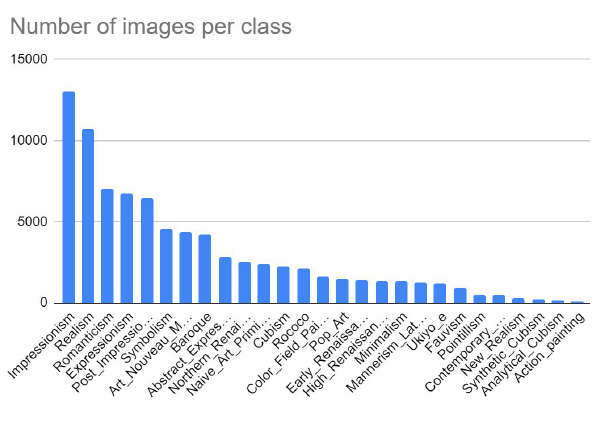}
    \caption{Distributions of number of images available for each of the 27 styles within the WikiArt dataset.}
    \label{fig:image_per_class}
\end{figure}

The dataset here is high class imbalanced which can be seen from the class-wise image distribution in Figure~\ref{fig:image_per_class}. To handle the problem of class imbalance, different data augmentation techniques have been used such as spacial transformations (Projective, affine, similarity, and Euclidean transformations) and non-spatial transformations (color, contrast, brightness, and sharpening transformations). The usages of these augmentations are explained clearly in the next section.

\section{\bf{METHODOLOGY}}
\label{sec:methodology}

Self-supervised representations of images play a crucial role in exploring the data variations across various transformations. In Supervised approaches, to handle class imbalance problem in various domains such as Medical Diagnosis, Disease Prediction and Art Style recognition, vanilla data augmentation techniques are applied to labeled data. Unlike these approaches, we can instead weakly train a semi-supervised model without any major reliance on the labeled datasets. 

We use the Ensemble of Auto-Encoding Transformations (EnAET) developed by Wang et al., 2019, to employ unsupervised/weakly supervised data augmentation techniques to explore various spatial and non-spatial transformations and their effects on the unlabeled data. On verifying the model performance across different benchmarks it is evident that self-supervised representations learned from an ensemble of transformations can enact a crucial role in significantly enhancing the performance of semi-supervised learning models.

The framework to recognize rare and exotic art styles is summarized as follows:
\begin{itemize}
    \item To train a semi-supervised model, an ensemble of both spatial and non-spatial transformations from both labeled and unlabeled data are used in a self-supervised setting.
    \item These set of AutoEncoding transformations are used as a regularization network by learning robust features across different image transformations and further improve the consistency of label predictions for transformed images t\textsubscript{k}(x) by minimizing their KL divergence with the original images x.
\end{itemize}

\subsection{\bf{Model}}

In the Semi-supervised learning paradigm, instead of pre-training the model, the proposed method devises network of AETs as a regularizer which compliments the Semi-supervised learning loss to train classifiers \cite{c7}.

The approach further enforces two aspects to enrich the final semi-supervised label predictions by accomplishing the following criteria:

\begin{itemize}
    \item Consistent Predictions: To maximize the prediction consistency even when the classification boundary to confidently predict labels is overlapping. The Mean Teacher model is used where weights of a teacher model is updated with an exponential moving average of the weights from all the student models. 
        
        \begin{equation}
\Theta_{\tau}^{\prime}=\alpha \Theta_{\tau-1}^{\prime}+(1-\alpha) \Theta_{\tau}
\end{equation}

    \item Confident Predictions: To further improve the boundary between classes to achieve highly confident label predictions MixUp\cite{c7} is employed to train a model with the linear combination of the inputs and their corresponding outputs.
\end{itemize}

Due to different applied transformations and augmentations, there could be a difference between the features extracted from original and transformed images. Thus, the transformation decoders D\textsubscript{k} could recover the corresponding transformations as long as the encoded features from the encoders capture the intricate details of fine visual structures in artistic images. The AutoEncoding Transformations can self-train a good feature representation which could be used along with the encoded features by an efficient semi-supervised classifier to explore an ensemble of spatial and non-spatial transformations.

   \begin{figure}[htbp]
      \centering
    
      \includegraphics[scale=0.4]{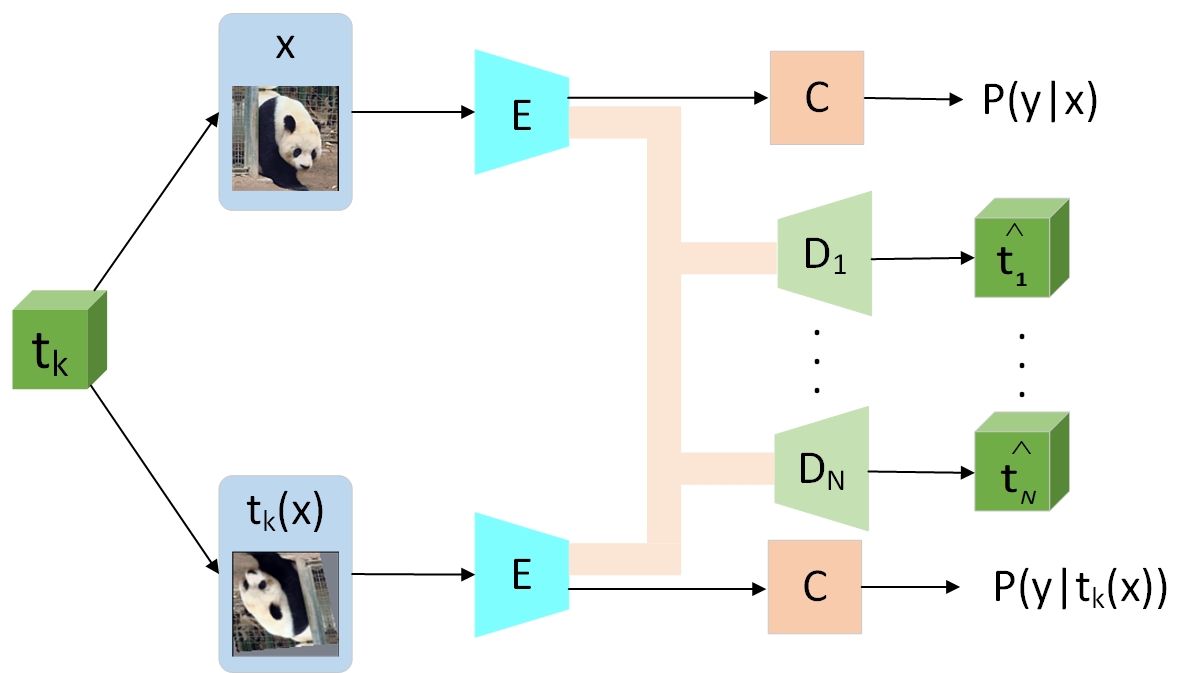}
      \caption{Architecture diagram for EnAET model (Wang et al., 2019)}
      \label{enaet}
      
   \end{figure}

\subsection{\bf{Framework}}

For each and every image x, five different transformations are applied, namely: Projective, Affine, Similarity, Euclidean  a combination of Color + Contrast + Brightness + Sharpness.

As shown in Figure \ref{enaet}, the model is split into three parts: an Encoder E, Classifier C, and a collection of Decoders D\textsubscript{k} for different transformations. The original input image x and all of its corresponding transformations t\textsubscript{k}(x) are loaded to respective encoders. Both the Encoders and Classifiers are arranged in a Siamese configuration (shared model weights).

The framework uses Wide ResNet-28-2 network which consists of four blocks. Among which the last block is used as Classifier C, while the Encoder E constitutes the other three blocks of the network. Similarly, all the Decoders D\textsubscript{k} use the same network architecture as that of the Classifier. But, the decoders are not configured in a Siamese configuration. The encoded representations of both original and transformed images are concatenated together and are fed to the decoders D\textsubscript{k}. Each decoder predicts the parameters of corresponding transformation t\textsubscript{k}. Finally, the classifiers C use the encoded representations from the encoder and generates appropriate label predictions for original and transformed images.

\subsection{\bf{Loss Functions}}

\begin{itemize}
    \item Model Loss: The idea is to minimize a linear combination of SSL and AET loss to train a classifier over the network weights theta.
    
    \begin{equation}
    \min _{\Theta} \mathcal{L}_{S S L}+\sum_{k=1}^{N} \lambda_{k} \mathcal{L}_{A E T_{k}}
    \end{equation}

    \item AutoEncoding Transformation Loss: Compute the Mean-Squared Error between the predicted transformation and the sampled transformation.
    
    \begin{equation}
    \mathcal{L}_{A E T_{k}}=\mathbb{E}_{x, t_{k}}\left\|D\left[E(x), E\left(t_{k}(x)\right)\right]-t_{k}\right\|^{2}
    \end{equation}
    
    \item Kullback-Leibler Divergence Loss: To make persistent predictions across different transformations, the approach is to minimize KL divergence between the predicted label on an original image x and predicted label on a transformed image t(x).
    
    \begin{equation}
    \mathcal{L}_{K L}=\mathbb{E}_{x, t} \sum_{y} P(y \mid x) \log \frac{P(y \mid x)}{P_{t}(y \mid x)}
    \end{equation}
    
    \item Semi-supervised Loss: Any loss function which could be used to train a semi-supervised classifier can be utilized. But, in this framework, MixMatch\cite{c7} Loss is employed.
    
    \begin{equation}
\left\{\begin{array}{l}
\mathcal{L}_{\mathcal{X}^{\prime}}=\mathbb{E}_{(x, y) \in \mathcal{X}^{\prime}} H(y, f(x, \Theta)) \\
\mathcal{L}_{\mathcal{U}^{\prime}}=\mathbb{E}_{(u, q) \in \mathcal{U}^{\prime}}\|f(u, \Theta)-q\|^{2} \\
\mathcal{L}_{m i x}=\mathcal{L}_{\mathcal{X}^{\prime}}+\lambda_{\mathcal{U}^{\prime}} \mathcal{L}_{\mathcal{U}^{\prime}}
  
\end{array}\right.
\end{equation}

\end{itemize}
\hfill

\subsection{\bf{Training Hyperparameters}}

Exhaustive Grid Search is performed to obtain optimal values for the following hyperparameters: Epochs, Learning Rate 0 for ADAM optimizer (Backbone network), Learning Rate 1 for SGD optimizer (AET regularizer network), KL Lambda value to maintain consistency, Warm Lambda are the warm factor values for different transformations in AET network, Max Lambda are the hyperparameters for different transformations in AET network, Data Portion, Batch Size and Beta to maintain consistency in MixMatch \cite{c8}. 

The final hyperparameters that were utilized are 100 training epochs, LR of 0.002, LR1 of 0.1, Batch Size of 128 images, KL Lambda = 1.0, Warm Lambda = (10, 7.5, 5, 2, 0.5), Max Lambda = (1, 0.75, 0.5, 0.2, 0.05), Data Portion = 1 and Beta = 75. The restriction for using a larger batch-size was the amount of volatile memory available in a single P100 GPU.

\section{\bf{BASELINES}}
\label{sec:baselines}

In order to evaluate the results of the trained model, we implemented two baselines - ResNet50, and ResNet50 with data augmentation.
% Use this sample document as your LaTeX source file to create your document. Save this file as {\bf root.tex}. You have to make sure to use the cls file that came with this distribution. If you use a different style file, you cannot expect to get required margins. Note also that when you are creating your out PDF file, the source file is only part of the equation. \emph{Your \TeX\ $\rightarrow$ PDF filter determines the output file size. Even if you make all the specifications to output a letter file in the source - if you filter is set to produce A4, you will only get A4 output.}

% It is impossible to account for all possible situation, one would encounter using \TeX. If you are using multiple \TeX\ files you must make sure that the ``MAIN`` source file is called root.tex - this is particularly important if your conference is using PaperPlaza's built in \TeX\ to PDF conversion tool.

\subsection{\bf{ResNet50}}

The ResNet50 architecture contains a convolution layer with 64 7*7 kernels with stride 2 followed by max pooling with stride 2. Then there are 4 groups of blocks such that each block is a 3-layer bottleneck block. At the end, there is a fully connected layer with output neurons corresponding to the number of classes. Rectified Linear Unit (ReLU) is used as the activation function for all weight layers, except for the last layer that uses softmax regression. This forms a total of 50 convolutional layers. 

Of the 65,168 WikiArt training images available to us, we use 45,000 images as train set and 20,168 images as validation set to train the model. Input images are first resized to 224x224 to simplify computation and preserve overall image structure. We then zero center the images and normalize them. Though our WikiArt dataset contains 65,168 training images, half of the classes contain less than 1,300 images per class. Hence we chose to use a model pre-trained for object recognition on ImageNet. During the training phase, we fine-tuned the entire model on our WikiArt dataset to obtain task specific features.

% Text heads organize the topics on a relational, hierarchical basis. For example, the paper title is the primary text head because all subsequent material relates and elaborates on this one topic. If there are two or more sub-topics, the next level head (uppercase Roman numerals) should be used and, conversely, if there are not at least two sub-topics, then no subheads should be introduced. Styles named ``Heading 1'', ``Heading 2'', ``Heading 3'', and ``Heading 4'' are prescribed.

\subsection{\bf{ResNet50 with Data Augmentation}}

To compare the performance of EnAET which leverages a wide range of image transformations we retain the same network and training setup as above but augment the dataset with the following transformations - random horizontal flip, rotation, translation, scaling and color-jitter. 

Horizontal image flip happens with a probability of 0.5, rotation is randomly drawn between 0\textdegree \ to 90\textdegree, horizontal and vertical translations are drawn between 0 and image width and between 0 and image height respectively, scaling factor is drawn between 1 to 2. Color-jitter consists of randomly modifying contrast, brightness and saturation of the input image independently.

% \begin{table}[h]
% \caption{An Example of a Table}
% \label{table_example}
% \begin{center}
% \begin{tabular}{|c||c|}
% \hline
% One  Two\\
% \hline
% Three  Four\\
% \hline
% \end{tabular}
% \end{center}
% \end{table}

%   \begin{figure}[thpb]
%       \centering
%       \framebox{\parbox{3in}{We suggest that you use a text box to insert a graphic (which is ideally a 300 dpi TIFF or EPS file, with all fonts embedded) because, in an document, this method is somewhat more stable than directly inserting a picture.
% }}
%       %\includegraphics[scale=1.0]{figurefile}
%       \caption{Inductance of oscillation winding on amorphous
%       magnetic core versus DC bias magnetic field}
%       \label{figurelabel}
%   \end{figure}

% Figure Labels: Use 8 point Times New Roman for Figure labels. Use words rather than symbols or abbreviations when writing Figure axis labels to avoid confusing the reader. As an example, write the quantity ``Magnetization'', or ``Magnetization, M'', not just ``M''. If including units in the label, present them within parentheses. Do not label axes only with units. In the example, write ``Magnetization (A/m)'' or ``Magnetization {A[m(1)]}'', not just ``A/m''. Do not label axes with a ratio of quantities and units. For example, write ``Temperature (K)'', not ``Temperature/K.''

\section{\bf{EXPERIMENTS AND RESULTS}}
\label{sec:exp_and_results}

In this section we aim to answer two scientific questions - \textit{\textbf{Q1:}} How well does EnAET perform on our dataset compared to the baselines and why?
\textit{\textbf{Q2:}} How much training data does EnAET require compared to the baselines?

\subsection{\bf{Performance Comparison and Analysis}}

\begin{figure}[htbp]
      \centering
    
      \includegraphics[scale=0.25]{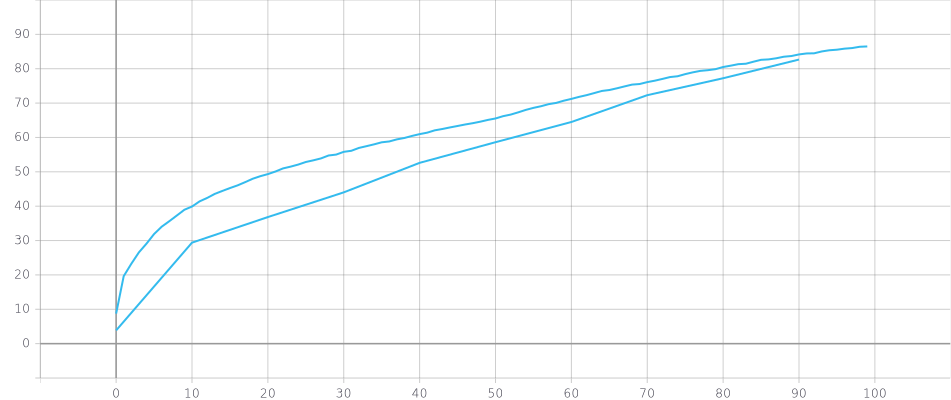}
      \caption{Test accuracies of EnAET from ephoch 1 to 100}
      \label{testacccurve}
   \end{figure}

To answer the first question we look at the overall classification accuracy of the three models on the test set in Table \ref{tab:table1}. The three models were evaluated on a test set consisting of 8616 images. We can see that the classification accuracy of the baseline ResNet50 model is the least at 50.1\% and with data augmentation the accuracy is increased slightly to 55\%. On the other hand, EnAET performs significantly better with an accuracy of 82.61\%. Observing the test accuracy curve for different epochs in Figure \ref{testacccurve}, we can see that EnAET starts to perform better than the baseline ResNet50 after epoch 40. It achieves the best accuracy scores after epoch 90.

\begin{figure}[htbp]
     
      \includegraphics[scale=0.37]{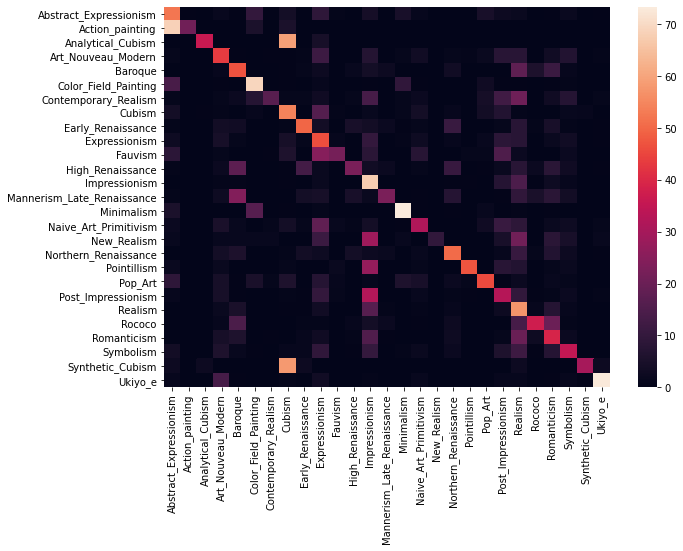}
      \caption{Confusion matrix plot of ResNet50. Y-axis specifies the true labels and x-axis specifies predicted labels.}
      \label{confmatrix}
   \end{figure}

To understand why EnAET performs so well in style classification, let us look at some drawbacks of the ResNet50 model performance. The confusion matrix of ResNet50 test predictions is presented in Figure \ref{confmatrix}. On examining the misclassifications in the vertical direction, it can be observed that images from many classes are often misclassified as the following five classes - Impressionism, Realism, Expressionism, Post Impressionism and Romanticism. Unsurprisingly, these are the most populated classes i.e, classes with the most number of training images. This means that precision of these highly populated classes are low as shown in Table \ref{tab:perclass}. On the other hand, on examining the least populated classes in the horizontal direction we can see that 70\% of Action painting images are wrongly predicted as Abstract expressionism. 60\% of Analytical cubism and Synthetic cubism images are falsely predicted as its bigger cousin Cubism. 30\% of New Realism images are misclassified as Impressionism and 20\% of them are misclassified as Realism. 30\% of Contemporary Realism images are misclassified as Realism. This means that recall of these less populated classes are low as shown in Table \ref{tab:perclass}. These classes have less than 500 images each in the train set. This shows that our imbalanced dataset is degrading the performance of the baseline ResNet. Even on augmenting our dataset, the accuracy only slightly improves but this problem persists.

Contrary to different Supervised methods which utilize various image augmentation strategies, EnAET overcomes the class imbalance problem by weakly training a semi-supervised model using meager amounts of labelled data and also employs unsupervised data augmentation techniques to explore various spatial and non-spatial transformations and their explicit effects on the unlabeled data. Further, crucial visual features in highly ambiguous classes are mastered by employing ensemble of AutoEncoding transformations as model regularizers.

\begin{table}[]
\caption{Comparison of test accuracies of EnAET with baselines}
\label{tab:table1}
\centering
\renewcommand{\arraystretch}{1.2}
\begin{tabular}{lll}
\hline
\multicolumn{1}{|l|}{\textbf{ResNet50}} & \multicolumn{1}{l|}{\textbf{\begin{tabular}[c]{@{}l@{}}ResNet50 with\\ data augmentation\end{tabular}}} & \multicolumn{1}{l|}{\textbf{EnAET}} \\ \hline
\multicolumn{1}{|l|}{50.1\%}            & \multicolumn{1}{l|}{55\%}                                                                               & \multicolumn{1}{l|}{82.61\%}        \\ \hline
                                        &                                                                                                         &                                    
\end{tabular}
\end{table}

\begin{table}[]
\begin{center}
\caption{Per-class precision and recall for the most populated classes and least populated classes. }
\label{tab:perclass}
\renewcommand{\arraystretch}{1.2}
\centering
\begin{tabular}{|l|l|l|ll}
\cline{1-3}
\textbf{Style}       & \textbf{Precision} & \textbf{Recall} &  &  \\ \cline{1-3}
Impressionism        & 34\%               & 68\%            &  &  \\ \cline{1-3}
Realism              & 28\%               & 57\%            &  &  \\ \cline{1-3}
Expressionism        & 30\%               & 46\%            &  &  \\ \cline{1-3}
Post Impressionism   & 34\%               & 33\%            &  &  \\ \cline{1-3}
Romanticism          & 34\%               & 40\%            &  &  \\ \cline{1-3}
Action Painting      & 66\%               & 21\%            &  &  \\ \cline{1-3}
Analytical Cubism    & 73\%               & 36\%            &  &  \\ \cline{1-3}
Synthetic Cubism     & 62\%               & 30\%            &  &  \\ \cline{1-3}
New Realism          & 86\%               & 10\%            &  &  \\ \cline{1-3}
Contemporary Realism & 47\%               & 17\%            &  &  \\ \cline{1-3}
\end{tabular}
\end{center}
\end{table}

\subsection{\bf{Experiment with data portions}}

To answer the second question of how much data the EnAET model requires to produce good results, we first note that the EnAET model is trained on 21,567 images which is only 33\% of the data used for the baseline ResNets.

We then experimented by using 5,391 training images which is only 8\% of the data used for training the baselines. In Table \ref{tab:dataportions}, the model trained on this train set is compared with other models. We can see that with just about 8\% of total training images used for the baselines, EnAET’s performance is comparable to the baseline ResNet50 though it is not better. Therefore, EnAET requires very little training data i.e. between 10-33\% of the data required for ResNet50.

\begin{table}[!hbt]
\begin{center}
\centering
\caption{Test accuracies of EnAET with different training data portions compared with baselines.}
\label{tab:dataportions}
\renewcommand{\arraystretch}{1.2}
\begin{tabular}{|l|l|l|}
\hline
\textbf{Model}                                                            & \textbf{Test Accuracy} & \textbf{\begin{tabular}[c]{@{}l@{}}\% of training \\ data used\end{tabular}} \\ \hline
ResNet50                                                                  & 50.1\%                 & 100\%                                                                        \\ \hline
\begin{tabular}[c]{@{}l@{}}ResNet50 with\\ data augmentation \end{tabular} & 55\%                   & 100\%                                                                        \\ \hline
EnAET                                                                     & 45.63\%                & 8\%                                                                          \\ \hline
EnAET                                                                     & \textbf{82.61\%}                & \textbf{33\%}                                                                         \\ \hline

\end{tabular}
\end{center}
\end{table}

\section{\bf{CONCLUSION}}
\label{sec:conclusion}

We clearly observe that style classification performance of the self-supervised EnAET model is better than the baseline methods. Our approach successfully overcomes the problems arising with highly imbalanced classes. It also requires only a fraction (10 - 33\%) of the training data required for the baseline models.

As future work, deeper encoder networks could be used to better represent image data into feature vectors. Additionally, distributed batch norm can be utilized to fasten the long model training routines. As alternative approaches we could explore contrastive learning methods which could be useful for high resolution image datasets similar to WikiArt and one-shot/few-shot approaches since we have very few images in several classes.

\addtolength{\textheight}{-12cm}   % This command serves to balance the column lengths

\printbibliography

@article{c1,
  author    = {Corneliu Florea and
               Razvan George Condorovici and
               Constantin Vertan and
               Raluca Boia and
               Laura Florea and
               Ruxandra Vr{\^{a}}nceanu},
  title     = {Pandora: Description of a Painting Database for Art Movement Recognition
               with Baselines and Perspectives},
  journal   = {CoRR},
  volume    = {abs/1602.08855},
  year      = {2016},
  url       = {http://arxiv.org/abs/1602.08855},
  archivePrefix = {arXiv},
  eprint    = {1602.08855},
  bibsource = {dblp computer science bibliography, https://dblp.org}
}

@inproceedings{c2,
  author    = {Gaowen Liu and
               Yan Yan and
               Elisa Ricci and
               Yi Yang and
               Yahong Han and
               Stefan Winkler and
               Nicu Sebe},
  editor    = {Qiang Yang and
               Michael J. Wooldridge},
  title     = {Inferring Painting Style with Multi-Task Dictionary Learning},
  booktitle = {Proceedings of the Twenty-Fourth International Joint Conference on
               Artificial Intelligence, {IJCAI} 2015, Buenos Aires, Argentina, July
               25-31, 2015},
  pages     = {2162--2168},
  publisher = {{AAAI} Press},
  year      = {2015},
  url       = {http://ijcai.org/Abstract/15/306},
  bibsource = {dblp computer science bibliography, https://dblp.org}
}

@article{c3,
  author    = {Babak Saleh and
               Ahmed M. Elgammal},
  title     = {Large-scale Classification of Fine-Art Paintings: Learning The Right
               Metric on The Right Feature},
  journal   = {CoRR},
  volume    = {abs/1505.00855},
  year      = {2015},
  url       = {http://arxiv.org/abs/1505.00855},
  archivePrefix = {arXiv},
  eprint    = {1505.00855},
  bibsource = {dblp computer science bibliography, https://dblp.org}
}

@inproceedings{c4,
  author    = {Wei Ren Tan and
               Chee Seng Chan and
               Hern{\'{a}}n E. Aguirre and
               Kiyoshi Tanaka},
  title     = {Ceci n'est pas une pipe: {A} deep convolutional network for fine-art
               paintings classification},
  booktitle = {2016 {IEEE} International Conference on Image Processing, {ICIP} 2016,
               Phoenix, AZ, USA, September 25-28, 2016},
  pages     = {3703--3707},
  publisher = {{IEEE}},
  year      = {2016},
  url       = {https://doi.org/10.1109/ICIP.2016.7533051},
  doi       = {10.1109/ICIP.2016.7533051},
  bibsource = {dblp computer science bibliography, https://dblp.org}
}

@inproceedings{c5,
  author    = {Adrian Lecoutre and
               Benjamin N{\'{e}}grevergne and
               Florian Yger},
  editor    = {Min{-}Ling Zhang and
               Yung{-}Kyun Noh},
  title     = {Recognizing Art Style Automatically in Painting with Deep Learning},
  booktitle = {Proceedings of The 9th Asian Conference on Machine Learning, {ACML}
               2017, Seoul, Korea, November 15-17, 2017},
  series    = {Proceedings of Machine Learning Research},
  volume    = {77},
  pages     = {327--342},
  publisher = {{PMLR}},
  year      = {2017},
  url       = {http://proceedings.mlr.press/v77/lecoutre17a.html},
  bibsource = {dblp computer science bibliography, https://dblp.org}
}

@article{c6,
  author    = {Kaiming He and
               Xiangyu Zhang and
               Shaoqing Ren and
               Jian Sun},
  title     = {Deep Residual Learning for Image Recognition},
  journal   = {CoRR},
  volume    = {abs/1512.03385},
  year      = {2015},
  url       = {http://arxiv.org/abs/1512.03385},
  archivePrefix = {arXiv},
  eprint    = {1512.03385},
  bibsource = {dblp computer science bibliography, https://dblp.org}
}

@article{c7,
  author    = {Xiao Wang and
               Daisuke Kihara and
               Jiebo Luo and
               Guo{-}Jun Qi},
  title     = {EnAET: Self-Trained Ensemble AutoEncoding Transformations for Semi-Supervised
               Learning},
  journal   = {CoRR},
  volume    = {abs/1911.09265},
  year      = {2019},
  url       = {http://arxiv.org/abs/1911.09265},
  archivePrefix = {arXiv},
  eprint    = {1911.09265},
  bibsource = {dblp computer science bibliography, https://dblp.org}
}

@article{c8,
  author    = {David Berthelot and
               Nicholas Carlini and
               Ian J. Goodfellow and
               Nicolas Papernot and
               Avital Oliver and
               Colin Raffel},
  title     = {MixMatch: {A} Holistic Approach to Semi-Supervised Learning},
  journal   = {CoRR},
  volume    = {abs/1905.02249},
  year      = {2019},
  url       = {http://arxiv.org/abs/1905.02249},
  archivePrefix = {arXiv},
  eprint    = {1905.02249},
  bibsource = {dblp computer science bibliography, https://dblp.org}
}

\end{document}